\documentclass{article}

\usepackage[final,nonatbib]{neurips_2022}
\usepackage[numbers]{natbib}

\usepackage[utf8]{inputenc} 
\usepackage[T1]{fontenc}    
\usepackage{hyperref}       
\usepackage{url}            
\usepackage{booktabs}       
\usepackage{amsfonts}       
\usepackage{nicefrac}       
\usepackage{microtype}      
\usepackage{xcolor}         
\usepackage{longtable}
\usepackage{graphicx}

\usepackage{xcolor}
\usepackage[export]{adjustbox}

\author{
  Vitali Petsiuk \\
  Boston University\\
  \texttt{vpetsiuk@bu.edu} \\
  \And
  Alexander Siemenn \\
  MIT \\
  \texttt{asiemenn@mit.edu} \\
  \And
  Saisamrit Surbehera \\
  Columbia University \\
  \texttt{ss6365@columbia.edu} \\
  \And
  Zad Chin \\
  Harvard University \\
  \texttt{zadchin@college.harvard.edu} \\
  \And
  Keith Tyser \\
  Boston University \\
  \texttt{ktyser@bu.edu} \\
  \And
  Gregory Hunter \\
  Columbia University \\
  \texttt{geh2129@columbia.edu} \\
  \And
  Arvind Raghavan \\
  Columbia University \\
  \texttt{ar4284@columbia.edu} \\
  \And
  Yann Hicke \\
  Cornell University \\
  \texttt{ylh8@cornell.edu} \\
  \And
  Bryan Plummer \\
  Boston University \\
  \texttt{bplum@bu.edu} \\
  \And
  Ori Kerret \\
  Ven Commerce \\
  \texttt{ori@ven.com} \\
  \And
  Tonio Buonassisi \\
  MIT \\
  \texttt{buonassi@mit.edu} \\
  \And
  Kate Saenko \\
  BU, MIT-IBM Watson AI Lab\\
  \texttt{saenko@bu.edu} \\
  \And
  Armando Solar-Lezama \\
  MIT \\
  \texttt{asolar@csail.mit.edu} \\
  \And
  Iddo Drori \\
  MIT, Columbia University, Boston University \\
  \texttt{idrori@csail.mit.edu, idrori@cs.columbia.edu, idrori@bu.edu} \\
  }

\title{Human Evaluation of Text-to-Image Models on a Multi-Task Benchmark}

\begin{document}

\maketitle

\begin{abstract}
We provide a new multi-task benchmark for evaluating text-to-image models. We perform a human evaluation comparing the most common open-source (Stable Diffusion) and commercial (DALL-E~2) models. Twenty computer science AI graduate students evaluated the two models, on three tasks, at three difficulty levels, across ten prompts each, providing 3,600 ratings. Text-to-image generation has seen rapid progress to the point that many recent models have demonstrated their ability to create realistic high-resolution images for various prompts. However, current text-to-image methods and the broader body of research in vision-language understanding still struggle with intricate text prompts that contain many objects with multiple attributes and relationships. We introduce a new text-to-image benchmark that contains a suite of thirty-two tasks over multiple applications that capture a model’s ability to handle different features of a text prompt. For example, asking a model to generate a varying number of the same object to measure its ability to count or providing a text prompt with several objects that each have a different attribute to identify its ability to match objects and attributes correctly. Rather than subjectively evaluating text-to-image results on a set of prompts, our new multi-task benchmark consists of challenge tasks at three difficulty levels (easy, medium, and hard) and human ratings for each generated image. 
\end{abstract}

\section{Introduction}
Spurred by large-scale pretraining on billions of image-text pairs, vision-language models have seen rapid progress in recent years. Large-scale models like CLIP~\cite{radford2021learning} and Flamingo~\cite{alayrac2022flamingo} have reported remarkable performance on dozens of benchmarks using a single model, even when using few or no task-specific training samples. Generating high-resolution images given a text prompt has improved in quality to such an extent with models like Stable Diffusion \cite{rombach2021highresolution}, Imagen~\cite{saharia2022photorealistic}, and DALL-E~2~\cite{ramesh2022hierarchical} that their influence has affected popular culture as illustrated in their use to generate magazine covers\footnote{\url{https://www.cosmopolitan.com/lifestyle/a40314356}}. 
There has been much recent progress in improving text-to-image models, allowing the synthesis of objects within novel contexts \cite{ruiz2022dreambooth} such as different backgrounds, illumination, and poses. However, these methods still need help generating images in complex scenes or where compositionality is essential. A critical bottleneck in further progress is the need for rigorous evaluation protocols, as current evaluation methods focus on prompts that do not fully account for the diverse settings these models must support~\cite{saharia2022photorealistic}.   

We propose a new text-to-image generation benchmark covering thirty-two tasks over multiple applications, each targeting a different capability of text-to-image generation models as shown in Table \ref{tab:50tasks}. For example, we may ask a model to produce varying numbers of an object to identify its ability to count or to generate an image with an object of a specified shape. We divide each task into three difficulty levels: easy, medium, and hard. For example, suppose the task is to synthesize different numbers of objects. In that case, the task may be easy: generating 1-3 objects, medium - generating 4-10 objects, and hard - generating more than ten objects. Next, we provide ten different instances for each task difficulty level. These instances are specific prompts that implement the tasks. We score text-to-image models on each of the thirty instances (ten for each of the three difficulty levels) for each of the fifty tasks and applications. Specifically, we run our benchmark on DALL-E~2~\cite{ramesh2022hierarchical}, and Stable Diffusion \cite{rombach2021highresolution}.

We can quantify and compare any new text-to-image generation model with our new benchmark. In this work, we perform a human evaluation of three tasks. Note, however, that many of the tasks may also be evaluated automatically, \textit{e.g.} by a neural network. 
For instance, incorporating spatial-aware methods ensures that prompts with spatial relationships and compositional elements are correctly generated. Using OCR mechanisms ensures that quoted text is legible and accurate.

Our key contributions are: (i) developing challenge tasks for state-of-the-art text-to-image generative models, (ii) defining human evaluation procedures 
, and (iii) Performing a human evaluation for a subset of tasks with 3,600 human ratings, comparing the performance of two of the most common open source (Stable Diffusion) and commercial (DALL-E~2) models.

\section{Related Work}
Text-to-image models may be roughly split into two types: autoregressive transformer-based models \cite{yu2022scaling} and diffusion-based models \cite{saharia2022photorealistic}. Prior state-of-the-art \cite{bar2022textlive,gal2022image,hertz2022prompt,ruiz2022dreambooth} handles specific limitations of text-to-image models, such as generating an image within the context or modifying object attributes automatically. A comprehensive and quantitative multi-task benchmark for text-to-image synthesis does not exist that covers a diverse set of tasks with varying difficulty levels. Our goal is to develop a benchmark that will become the gold standard in the field for evaluating text-to-image models that will endure time. 

Text-to-image models are commonly evaluated by the Inception Score (IS) and the Fréchet Inception Distance (FID). Both of these metrics are based on Inception v3 classifier. These measures, therefore, are designed for the unconditional setting and are primarily trained on single-object images. We have seen several approaches which rectify these shortcomings. Imagen~\cite{saharia2022photorealistic} introduced DrawBench, a benchmark with 11 categories with approximately 200 prompts total. Human raters (25 participants) were asked to choose a better set of generated images from two models regarding image fidelity and image-text alignment. Categories are: colors
counting, conflicting, DALL-E~2, description, misspellings, positional, rare words, Reddit, text.

DALL-Eval~\cite{Cho2022DallEval} proposed PaintSkills to test skills of the generative models --- specified object generation, counting, color, and spatial relations. It utilizes the Unity engine to test these tasks using predefined sets of objects, a subset of MS-COCO~\cite{lin2014coco} objects, colors, and spatial relations. We propose a more comprehensive benchmark of tasks at a finer level of detail, with three difficulty levels. Localized Narratives~\cite{pt2020localized} is a multi-modal image captioning approach that can be adapted to benchmarking images. Text captions are first generated by human annotators whose cursor movement and voice commentary hover their cursor over the image to provide richness and accuracy. PartiPrompts, a holistic benchmark of 1,600 English prompts~\cite{yu2022parti}, compared to Localized Narratives, is better in probing model capabilities on open-domain text-to-image generation. The 1,600 prompts span 12 different categories and 11 challenge aspects. The 12 categories are artifacts, animals, indoor scenes, produce and plants, abstract, arts, food \& beverage, vehicles, illustrations, outdoor scenes, people, and world knowledge, while the 11 challenge aspect are basic, fine-grained detail, properties \& positioning, linguistic structures, perspective, quantity, writing \& symbols, complex, imagination, style \& format and simple detail. The image quality and the alignment of the generated image with the input text are evaluated.

\def \imagewidth{18mm}
\def \colimgwidth{\dimexpr (\imagewidth)*2+6mm}
\def \coltextwidth{\dimexpr \linewidth-(\colimgwidth)*2}

\section{Methods}

We create a comprehensive multi-task text-to-image generation benchmark of thirty-two diverse tasks, as shown in Table \ref{tab:50tasks}. 
The motivation behind selecting benchmark tasks is to cover a wide range of downstream applications that benefit from high-quality text-to-image models, including:
(1) graphic design (generating new designs for websites), 
(2) e-Commerce (generating personalized ads), 
(3) architectural planning and design (generating new renderings of buildings, interior design, and creative home design), 
(4) real-estate listings (generating furnished versions of unfurnished apartment and house photos for advertisement), 
(5) education (generating personalized digital learning interfaces with customized enhancements), 
(6) user interface and user experience (generating design templates for mobile and desktop applications), 
(7) stop-motion video (generating frames for short animations), 
(8) cosmetics (generating realistic images showcasing products), 
(9) stock photos (generating large amounts of stock images for general audiences), 
(10) product design (quickly prototyping digital and physical products), 
(11) illustrations (generating professional artwork for custom purposes), 
(12) synthetic data (generating synthetic data for boosting training samples size),
(13) social media (generating memes and shareable content), 
(14) image recommendation (generating recommendations based on user preferences), 
(15) gaming (using natural language to create complex scenes for video games), 
(16) proteomics (designing new proteins visualizing existing structures),
(17) material science (designing new crystals).

We use the benchmark to compare different models, comparing Stable Diffusion \cite{rombach2021highresolution} \footnote{Publicly available \url{https://beta.dreamstudio.ai/}} and DALL-E~2~\cite{ramesh2022hierarchical}, and identify their limitations. Our evaluation protocol consists of human ratings between 1 (worst) and 5 (best) of tasks at three difficulty levels. 
The three difficulty levels are currently assigned heuristically based on the experience with failure modes of generative models. Examples of cases in which human evaluation is required are: (i) concepts that are difficult to define, such as successfully combining objects that are rarely co-occurring in the real world; (ii) complex tasks, such as images that require common sense; and (iii) cases where human expertise is essential such as generating images without racial or gender bias. We obtained 3,600 human ratings: twenty graduate students, two models, three difficulty levels, and ten prompts each. The images were all generated with identical default model parameters and evaluated on the same scale, scoring between 1 (worst) and 5 (best).

\begin{table*}[t!]
\small
    \centering
    \begin{tabular}{r|ll|ll|ll||ll}
    \toprule
    \textbf{Difficulty} & \multicolumn{2}{|c|}{\textbf{Easy}} & \multicolumn{2}{|c|}{\textbf{Medium}} & \multicolumn{2}{|c||}{\textbf{Hard}} & \multicolumn{2}{|c}{\textbf{Average}}\\
\textbf{Task / Model} & SD & DALL-E~2 & SD & DALL-E~2 & SD & DALL-E~2 & SD & DALL-E~2 \\
    \hline
    Counting & 74.8 & 91.8 & 52.2 & 51.4 & 36.1 & 54.0 & 54.4 & 65.7 \\
    \hline
    Faces & 72.5 & 93.5 & 74.0 & 74.3 & 64.2 & 77.2 & 70.2 & 81.7 \\
    \hline
    Shapes & 70.8 & 67.1 & 56.8 & 46.0 & 45.1 & 57.3 & 57.6 & 56.8\\
    \bottomrule
    \end{tabular}
    \caption{Normalized ratings (\%) for human evaluation of Stable Diffusion and DALL-E~2 across the three tasks of counting, shapes, and faces. DALL-E~2 outperforms Stable Diffusion on Counting and Faces tasks, and Stable Diffusion shows a minor advantage on the shapes task. On 6 out of 9 sub-tasks, DALL-E~2 produces better images.
}
\label{tab:human-eval-summary}
\end{table*}

\begin{table*}
\small
    \centering
    \begin{tabular}{l|l}
        \toprule
        \textbf{Id} & \textbf{Task}\\
        \hline
        1 & Generating a specified number of objects\\
        2 & Generating objects with specified spatial positioning\\
        3 & Combining objects very rarely co-occurring in the world\\
        4 & Generating images obeying physical rendering rules of shadows, reflections, and acoustics\\
        5 & Generating objects with specified colors\\
        6 & Generating conflicting interactions between objects\\
        7 & Understanding complex and long text prompts describing objects\\
        8 & Understanding misspelled prompts\\
        9 & Handling absurd requests\\
        10 & Understanding rare words\\
        11 & Incorporating quoted text with correct spelling\\
        12 & Understanding negation and counter-examples\\
        13 & Understanding anaphora and phrases that refer to other parts of the prompt\\
        14 & Aligning text as specified in the prompt\\
        15 & Generating common-sense images\\
        16 & Removing objects without needing manual annotation\\
        17 & Removing content that is not child-safe\\
        18 & Editing the color of objects without marking them manually\\
        19 & Replacing objects without marking them manually\\
        20 & Generating objects with abstract shapes\\
        21 & Objects obeying physics rules\\
        22 & Generating images without racial or gender bias\\
        23 & Understanding comparative concepts like fewer and more\\
        24 & Photo-realistic faces\\
        25 & Understanding prompts regarding weather\\
        26 & Handling multi-lingual prompts\\
        27 & Duplicating objects perfectly\\
        28 & Generating multiple camera viewpoints of the same scene\\
        29 & Generating realistic faces with a specific emotion\\
        30 & Generating well-known faces\\
        31 & Generating a thumbnail summary for text and video\\
        32 & Changing dimensions of an image without losing information\\
        \bottomrule
    \end{tabular}
    \caption{Multi-task text-to-image benchmark and applications: we propose a series of tasks on which text-to-image models could be evaluated. The tasks are selected to cover a multitude of downstream applications.}
    \label{tab:50tasks}
\end{table*}

\section{Results}

We survey twenty students that evaluate the performance of Stable Diffusion and DALL-E~2. Each student evaluated the three tasks at three levels of difficulty and ten prompts each, providing a rating between 1 (worst) and 5 (best). We collected a total of 3,600 scores.
The results are summarized in Table~\ref{tab:human-eval-summary}, and a detailed breakdown of the evaluations by prompts are available in Table~\ref{tab:human-eval-detail} of the Appendix. Our human evaluation shows that DALL-E~2 (65.7\%) performs better on the counting task than Stable Diffusion (54.4\%). On the Faces tasks, both models perform very well, and DALL-E~2 (81.7\%) performs better than Stable Diffusion (70.2\%), and on the Shapes task, both models perform equally well (56.8\% compared to 57.6\%). Performance gracefully degrades as the tasks are more difficult, except for DALL-E~2, which performs slightly better on the hard than on the medium Shapes task.

\section{Conclusion}
We propose a new quantitative way of evaluating text-to-image generative models using a benchmark covering many model competencies and applications. Initial human evaluation on a subset of benchmark tasks shows a slight advantage of DALL-E~2 over Stable Diffusion. Our proposed benchmark allows for testing individual competencies and limitations of the different generative models. Understanding the limitations is critical for picking the suitable model for each task and application, advancing the quality of generative models, and aligning their performance with human goals.

\bibliographystyle{plainnat}
\bibliography{bibliography}

\begin{thebibliography}{14}
\providecommand{\natexlab}[1]{#1}
\providecommand{\url}[1]{\texttt{#1}}
\expandafter\ifx\csname urlstyle\endcsname\relax
  \providecommand{\doi}[1]{doi: #1}\else
  \providecommand{\doi}{doi: \begingroup \urlstyle{rm}\Url}\fi

\bibitem[Alayrac et~al.(2022)Alayrac, Donahue, Luc, Miech, Barr, Hasson, Lenc,
  Mensch, Millican, Reynolds, et~al.]{alayrac2022flamingo}
Jean-Baptiste Alayrac, Jeff Donahue, Pauline Luc, Antoine Miech, Iain Barr,
  Yana Hasson, Karel Lenc, Arthur Mensch, Katie Millican, Malcolm Reynolds,
  et~al.
\newblock Flamingo: a visual language model for few-shot learning.
\newblock \emph{arXiv preprint arXiv:2204.14198}, 2022.

\bibitem[Bar-Tal et~al.(2022)Bar-Tal, Ofri-Amar, Fridman, Kasten, and
  Dekel]{bar2022textlive}
Omer Bar-Tal, Dolev Ofri-Amar, Rafail Fridman, Yoni Kasten, and Tali Dekel.
\newblock Text2live: Text-driven layered image and video editing.
\newblock \emph{arXiv preprint arXiv:2204.02491}, 2022.

\bibitem[Cho et~al.(2022)Cho, Zala, and Bansal]{Cho2022DallEval}
Jaemin Cho, Abhay Zala, and Mohit Bansal.
\newblock Dall-eval: Probing the reasoning skills and social biases of
  text-to-image generative transformers.
\newblock \emph{arXiv preprint arXiv:2202.04053}, 2022.

\bibitem[Gal et~al.(2022)Gal, Alaluf, Atzmon, Patashnik, Bermano, Chechik, and
  Cohen-Or]{gal2022image}
Rinon Gal, Yuval Alaluf, Yuval Atzmon, Or~Patashnik, Amit~H Bermano, Gal
  Chechik, and Daniel Cohen-Or.
\newblock An image is worth one word: Personalizing text-to-image generation
  using textual inversion.
\newblock \emph{arXiv preprint arXiv:2208.01618}, 2022.

\bibitem[Hertz et~al.(2022)Hertz, Mokady, Tenenbaum, Aberman, Pritch, and
  Cohen-Or]{hertz2022prompt}
Amir Hertz, Ron Mokady, Jay Tenenbaum, Kfir Aberman, Yael Pritch, and Daniel
  Cohen-Or.
\newblock Prompt-to-prompt image editing with cross attention control.
\newblock \emph{arXiv preprint arXiv:2208.01626}, 2022.

\bibitem[Lin et~al.(2014)Lin, Maire, Belongie, Hays, Perona, Ramanan,
  Doll{\'a}r, and Zitnick]{lin2014coco}
Tsung-Yi Lin, Michael Maire, Serge Belongie, James Hays, Pietro Perona, Deva
  Ramanan, Piotr Doll{\'a}r, and C~Lawrence Zitnick.
\newblock Microsoft coco: Common objects in context.
\newblock In \emph{European conference on computer vision}, pages 740--755.
  Springer, 2014.

\bibitem[Pont-Tuset et~al.(2020)Pont-Tuset, Uijlings, Changpinyo, Soricut, and
  Ferrari]{pt2020localized}
Jordi Pont-Tuset, Jasper Uijlings, Soravit Changpinyo, Radu Soricut, and
  Vittorio Ferrari.
\newblock Connecting vision and language with localized narratives.
\newblock In \emph{European conference on computer vision}, pages 647--664.
  Springer, 2020.

\bibitem[Radford et~al.(2021)Radford, Kim, Hallacy, Ramesh, Goh, Agarwal,
  Sastry, Askell, Mishkin, Clark, et~al.]{radford2021learning}
Alec Radford, Jong~Wook Kim, Chris Hallacy, Aditya Ramesh, Gabriel Goh,
  Sandhini Agarwal, Girish Sastry, Amanda Askell, Pamela Mishkin, Jack Clark,
  et~al.
\newblock Learning transferable visual models from natural language
  supervision.
\newblock In \emph{International Conference on Machine Learning}, pages
  8748--8763. PMLR, 2021.

\bibitem[Ramesh et~al.(2022)Ramesh, Dhariwal, Nichol, Chu, and
  Chen]{ramesh2022hierarchical}
Aditya Ramesh, Prafulla Dhariwal, Alex Nichol, Casey Chu, and Mark Chen.
\newblock Hierarchical text-conditional image generation with clip latents.
\newblock \emph{arXiv preprint arXiv:2204.06125}, 2022.

\bibitem[Rombach et~al.(2022)Rombach, Blattmann, Lorenz, Esser, and
  Ommer]{rombach2021highresolution}
Robin Rombach, Andreas Blattmann, Dominik Lorenz, Patrick Esser, and Bj{\"o}rn
  Ommer.
\newblock High-resolution image synthesis with latent diffusion models.
\newblock In \emph{Proceedings of the IEEE/CVF Conference on Computer Vision
  and Pattern Recognition}, pages 10684--10695, 2022.

\bibitem[Ruiz et~al.(2022)Ruiz, Li, Jampani, Pritch, Rubinstein, and
  Aberman]{ruiz2022dreambooth}
Nataniel Ruiz, Yuanzhen Li, Varun Jampani, Yael Pritch, Michael Rubinstein, and
  Kfir Aberman.
\newblock {DreamBooth}: {F}ine tuning text-to-image diffusion models for
  subject-driven generation.
\newblock \emph{arXiv preprint arXiv:2208.12242}, 2022.

\bibitem[Saharia et~al.(2022)Saharia, Chan, Saxena, Li, Whang, Denton,
  Ghasemipour, Ayan, Mahdavi, Lopes, et~al.]{saharia2022photorealistic}
Chitwan Saharia, William Chan, Saurabh Saxena, Lala Li, Jay Whang, Emily
  Denton, Seyed Kamyar~Seyed Ghasemipour, Burcu~Karagol Ayan, S~Sara Mahdavi,
  Rapha~Gontijo Lopes, et~al.
\newblock Photorealistic text-to-image diffusion models with deep language
  understanding.
\newblock \emph{arXiv preprint arXiv:2205.11487}, 2022.

\bibitem[Yu et~al.(2022{\natexlab{a}})Yu, Xu, Koh, Luong, Baid, Wang,
  Vasudevan, Ku, Yang, Ayan, et~al.]{yu2022parti}
Jiahui Yu, Yuanzhong Xu, Jing~Yu Koh, Thang Luong, Gunjan Baid, Zirui Wang,
  Vijay Vasudevan, Alexander Ku, Yinfei Yang, Burcu~Karagol Ayan, et~al.
\newblock Scaling autoregressive models for content-rich text-to-image
  generation.
\newblock \emph{arXiv preprint arXiv:2206.10789}, 2022{\natexlab{a}}.

\bibitem[Yu et~al.(2022{\natexlab{b}})Yu, Xu, Koh, Luong, Baid, Wang,
  Vasudevan, Ku, Yang, Ayan, et~al.]{yu2022scaling}
Jiahui Yu, Yuanzhong Xu, Jing~Yu Koh, Thang Luong, Gunjan Baid, Zirui Wang,
  Vijay Vasudevan, Alexander Ku, Yinfei Yang, Burcu~Karagol Ayan, et~al.
\newblock Scaling autoregressive models for content-rich text-to-image
  generation.
\newblock \emph{arXiv preprint arXiv:2206.10789}, 2022{\natexlab{b}}.

\end{thebibliography}

\newpage
\appendix
\section{Appendix}

\begin{table*}[h!]
\small
\centering
\begin{tabular}{r|cc|cc|cc}
\toprule
\textbf{Difficulty} & \multicolumn{2}{|c|}{\textbf{Easy}} & \multicolumn{2}{|c|}{\textbf{Medium}} & \multicolumn{2}{|c}{\textbf{Hard}}\\
\textbf{Prompt / Model} & SD & DALL-E~2 & SD & DALL-E~2 & SD & DALL-E~2\\
\hline
1  & 96 & 100 & 69 & 32 & 30 & 57 \\
2  & 93 & 97  & 62 & 62 & 28 & 52 \\
3  & 49 & 97  & 34 & 36 & 22 & 61 \\
4  & 87 & 99  & 38 & 46 & 37 & 44 \\
5  & 51 & 90  & 77 & 64 & 46 & 57 \\
6  & 97 & 50  & 54 & 45 & 35 & 82 \\
7  & 92 & 93  & 51 & 79 & 47 & 48 \\
8  & 67 & 97  & 38 & 39 & 37 & 45 \\
9  & 43 & 98  & 43 & 50 & 30 & 55 \\
10 & 73 & 97  & 56 & 61 & 49 & 39 \\
\hline
1  & 69 & 96 & 79 & 86 & 66 & 73 \\
2  & 82 & 93 & 83 & 95 & 70 & 69 \\
3  & 76 & 96 & 76 & 75 & 68 & 80 \\
4  & 97 & 98 & 70 & 87 & 91 & 91 \\
5  & 23 & 92 & 81 & 86 & 58 & 90 \\
6  & 75 & 96 & 71 & 77 & 74 & 77 \\
7  & 60 & 92 & 91 & 20 & 53 & 95 \\
8  & 58 & 84 & 76 & 66 & 41 & 81 \\
9  & 96 & 94 & 74 & 77 & 59 & 51 \\
10 & 89 & 94 & 39 & 74 & 62 & 65 \\
\hline
1  & 93 & 92 & 73 & 30 & 36 & 41 \\
2  & 94 & 84 & 49 & 36 & 67 & 64 \\
3  & 89 & 77 & 32 & 32 & 44 & 62 \\
4  & 85 & 49 & 48 & 38 & 44 & 50 \\
5  & 49 & 38 & 69 & 42 & 30 & 66 \\
6  & 89 & 45 & 45 & 83 & 47 & 71 \\
7  & 52 & 80 & 65 & 52 & 34 & 70 \\
8  & 37 & 32 & 32 & 42 & 49 & 47 \\
9  & 79 & 90 & 84 & 37 & 51 & 37 \\
10 & 41 & 84 & 71 & 68 & 49 & 65 \\
\hline
\hline
Average: & 72.70 & 84.13 & 61.00 & 57.23 & 48.47 & 62.83\\
\bottomrule
\end{tabular}
\caption{Normalized ratings (\%) for human evaluation of Stable Diffusion and DALL-E~2 across the three tasks of counting, shapes, and faces for each of the ten prompts.} 
\label{tab:human-eval-detail}
\end{table*}

\begin{figure*}[ht]
    \centering
    \includegraphics[width=0.925\linewidth]{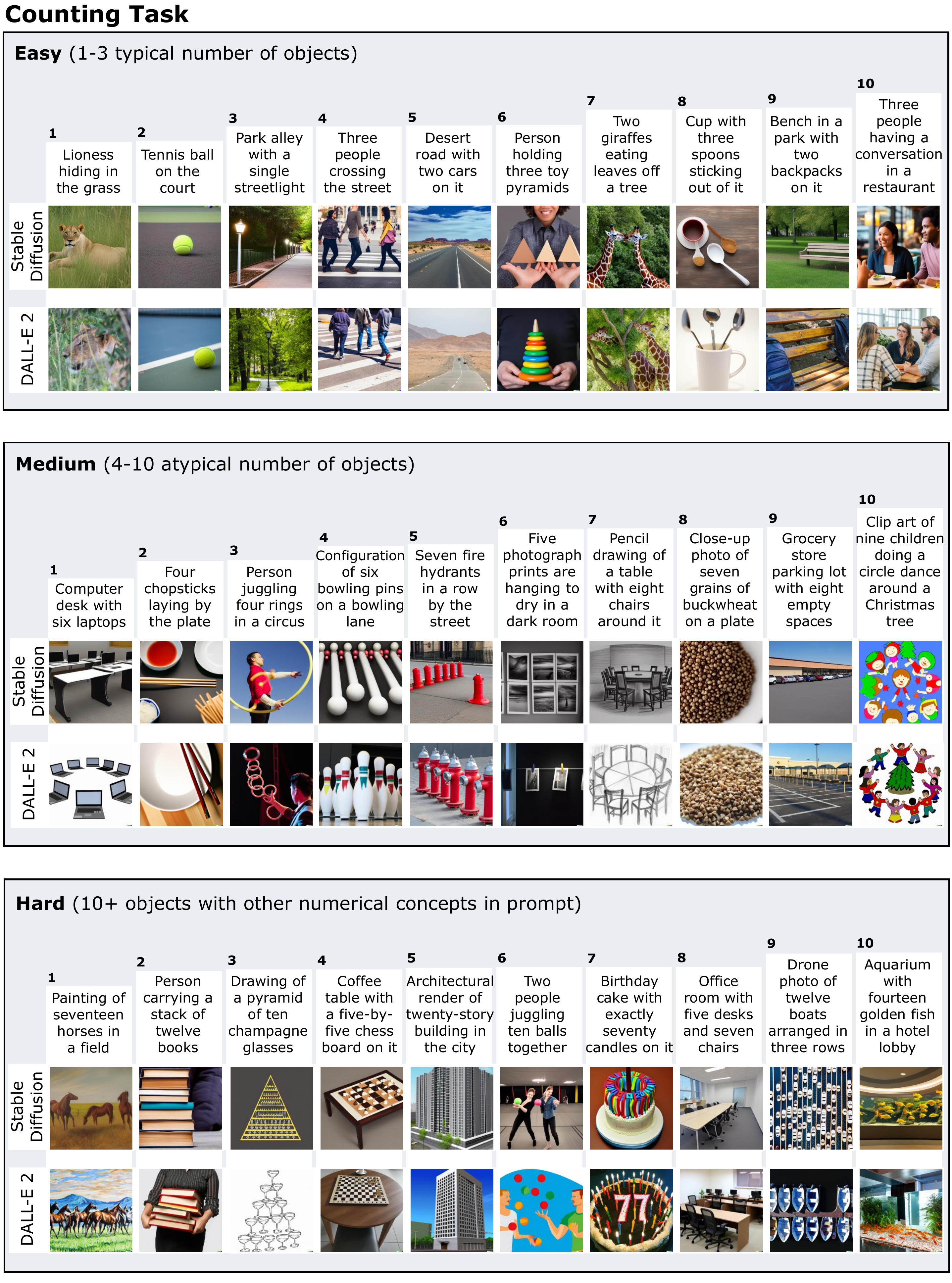}
    \caption{Evaluation of image generation on a counting task at various difficulties. Each panel contains tasks with different difficulties. The columns correspond to the text prompt, and the rows correspond to the model used to generate the image: (i) Stable Diffusion and (ii) DALL-E~2.
    The prompts used to generate the images are classified into three different difficulties: (i) easy difficulty tasks consisting of generating 1--3 objects, \textit{e.g., two cars, three people}; (ii) medium difficulty tasks consisting of generating 4--10 objects, including uncommon combinations of quantities and objects, \textit{e.g., six bowling pins, seven fire hydrants}; (iii) hard difficulty tasks consisting of 10 or more objects with other numerical concepts in the prompt, \textit{e.g., twelve boats arranged in three rows}.}
    \label{fig:counting}
\end{figure*}

\begin{figure*}[ht]
    \centering
    \includegraphics[width=0.925\linewidth]{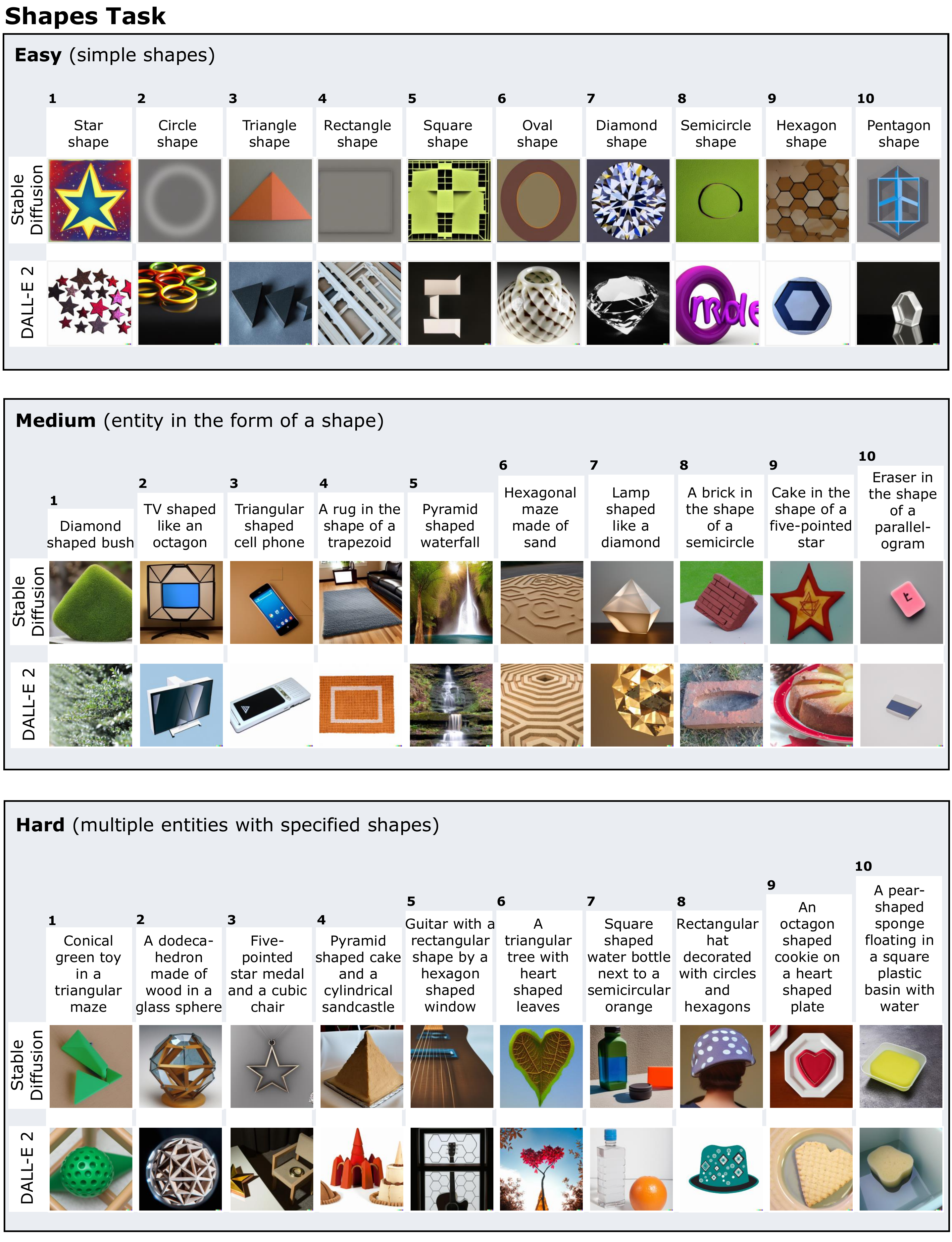}
    \caption{Evaluation of image generation on a shape task at various difficulties. Each panel contains tasks at different difficulties where the columns correspond to the text prompt, and the rows correspond to the model used to generate the image: (i) Stable Diffusion and (ii) DALL-E~2
    The prompts used to generate the images are classified into three different difficulties: (i) easy difficulty tasks consisting of generating simple shapes, \textit{e.g., circle, star}; (ii) medium difficulty tasks consisting of generating entities in the form of a shape, \textit{e.g., hexagonal maze, octagonal TV}; (iii) hard difficulty tasks consisting of multiple entities, each of a specified shape, \textit{e.g., square-shaped water bottle next to a semicircular orange}.}
    \label{fig:shapes}
\end{figure*}

\begin{figure*}[ht]
    \centering
    \includegraphics[width=0.925\linewidth]{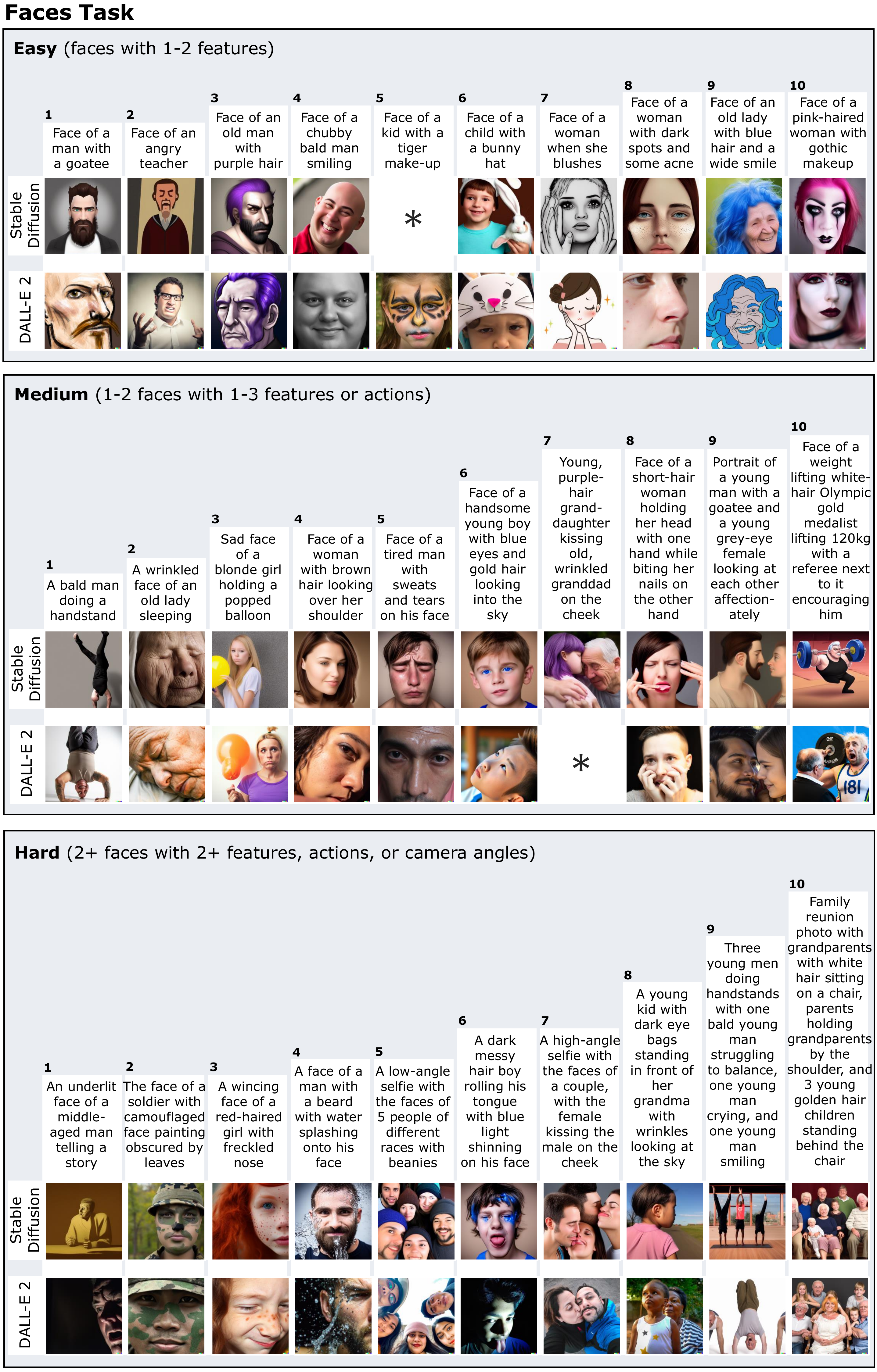}
    \caption{The prompts used to generate the images are classified into three different difficulties: generating photo-realistic faces given (i) easy: 1 to 2 features for an individual; (ii) medium: different 1 to 3 features for each individual in a group of 1 to 2 people with easy angle and posture; (iii) hard: given more than two features for each individual in a group of more than two people with challenging angle, posture, lighting or occlusion. Asterisk $(*)$ indicates a failure to generate an image because of the model's content filters.}
    \label{fig:faces}
\end{figure*}

\end{document}